\documentclass{ceurart}
\usepackage{booktabs}

\newcommand{\texasACC}{36.61}
\newcommand{\texasNMI}{8.08}
\newcommand{\wiscACC}{35.14}
\newcommand{\wiscNMI}{9.72}
\newcommand{\chamACC}{31.07}
\newcommand{\chamNMI}{8.80}

\begin{document}



\copyrightyear{2026}
\copyrightclause{Copyright for this paper by its authors. Use permitted under Creative Commons License Attribution 4.0 International (CC BY 4.0).}
\conference{ISWC 2026 Companion Volume, October 25--29, 2026, Bari, Italy}

\title{When Connected Does Not Mean Similar: Charting the Homophily
  Boundary of SNAP-KG for Streaming Entity Integration}

\author{Jui-Chien Lin}[orcid=0009-0000-7640-1140, email=linj26@rpi.edu]
\author{Oshani Seneviratne}[orcid=0000-0001-8518-917X, email=senevo@rpi.edu]
\address{Rensselaer Polytechnic Institute, Troy, NY 12180, USA}

\begin{abstract}
SNAP-KG is a framework for assigning newly arriving entities to semantic
communities in a growing knowledge graph (KG) using only their raw features,
with no graph access and no retraining at inference time. It was evaluated on five multi-view
benchmarks and a 2.4M-node OGB-WikiKG2 KG. In each of these datasets, at least one
graph view is \emph{homophilous}, meaning that connected nodes usually
belong to the same class, and SNAP-KG performs well on all of them. This paper asks what happens outside that setting. We extend the evaluation
to three heterophilous graphs (Texas, Wisconsin, Chameleon) and measure
the edge homophily of every view. When no homophilous view is available,
clustering quality drops sharply for both SNAP-KG and the transductive baselines used in its original evaluation.
What decides this is the homophily of the relation, not the number of relations. Multi-view fusion still helps, but only when at least one homophilous relation provides a reliable foundation.
The
homophily assumption is therefore shared by the whole method family, not
specific to SNAP-KG. We argue that heterophilous multi-view clustering is a separate research problem, outside the scope of this work. As future work, we outline how a heterophily-aware teacher could be distilled into SNAP-KG's projector to serve both homophilous and heterophilous KGs.
\end{abstract}



\maketitle

\section{Introduction}

KG construction pipelines must keep integrating newly arriving entities
into a growing graph. A new entity arrives with no graph connectivity:
it emerges from the upstream acquisition phase as a raw feature vector,
and it must be placed into a semantic community before entity resolution
(ER) and link prediction (LP) can run over a small candidate set instead
of the entire node population. Multi-view graph clustering methods build
high-quality, relationally informed communities by treating each KG
relation type as a separate structural view, but they are transductive:
they cannot place unseen entities without retraining~\cite{pan2021multi}.
Streaming Node Assignment via Projection for KG Entity
Integration (SNAP-KG)~\cite{lin2026snapkg} closes this gap.
SNAP-KG distills a multi-view Graph Neural Network (GNN)
and Transformer encoder into a lightweight Multi-Layer Perceptron (MLP)
projector $f_\phi$~\cite{zhang2021graph}. At inference, any new entity is
embedded from its raw features alone and assigned to the nearest cluster
centroid, with no graph access and no retraining. On five multi-view
benchmarks and a 2.4M-node OGB-WikiKG2 KG ~\cite{hu2020open}, this matches the quality of full
retraining at an order-of-magnitude lower cost.

SNAP-KG depends on one structural assumption:
\emph{connected nodes tend to be similar}, a property known as
homophily~\cite{zhu2020beyond}. All five benchmarks used to evaluate
SNAP-KG satisfy this property. This raises a natural question:
what happens when it does not hold?

\paragraph{Contribution.}
This paper extends the evaluation of SNAP-KG in four ways:
\textbf{(i)}~new clustering experiments on three standard heterophilous
graphs, where connected nodes tend to belong to \emph{different} classes
(Texas, Wisconsin, Chameleon)~\cite{pei2020geom,rozemberczki2021multi},
run over five seeds; \textbf{(ii)}~a measurement of per-view edge
homophily for all eight evaluated datasets, which links SNAP-KG's
clustering quality to a graph property that can be checked before
deployment, together with a single-view comparison that separates
heterophily from the number of views; \textbf{(iii)}~evidence that the transductive baselines from
the paper's evaluation drop in the same way, set against the dedicated
heterophilous-clustering literature; and \textbf{(iv)}~a concrete
research roadmap for extending streaming entity integration to
heterophilous KGs.

\section{SNAP-KG and Its Homophily Assumption}
\label{sec:nutshell}

\begin{figure}[t]
\centering
\includegraphics[width=0.8\linewidth]{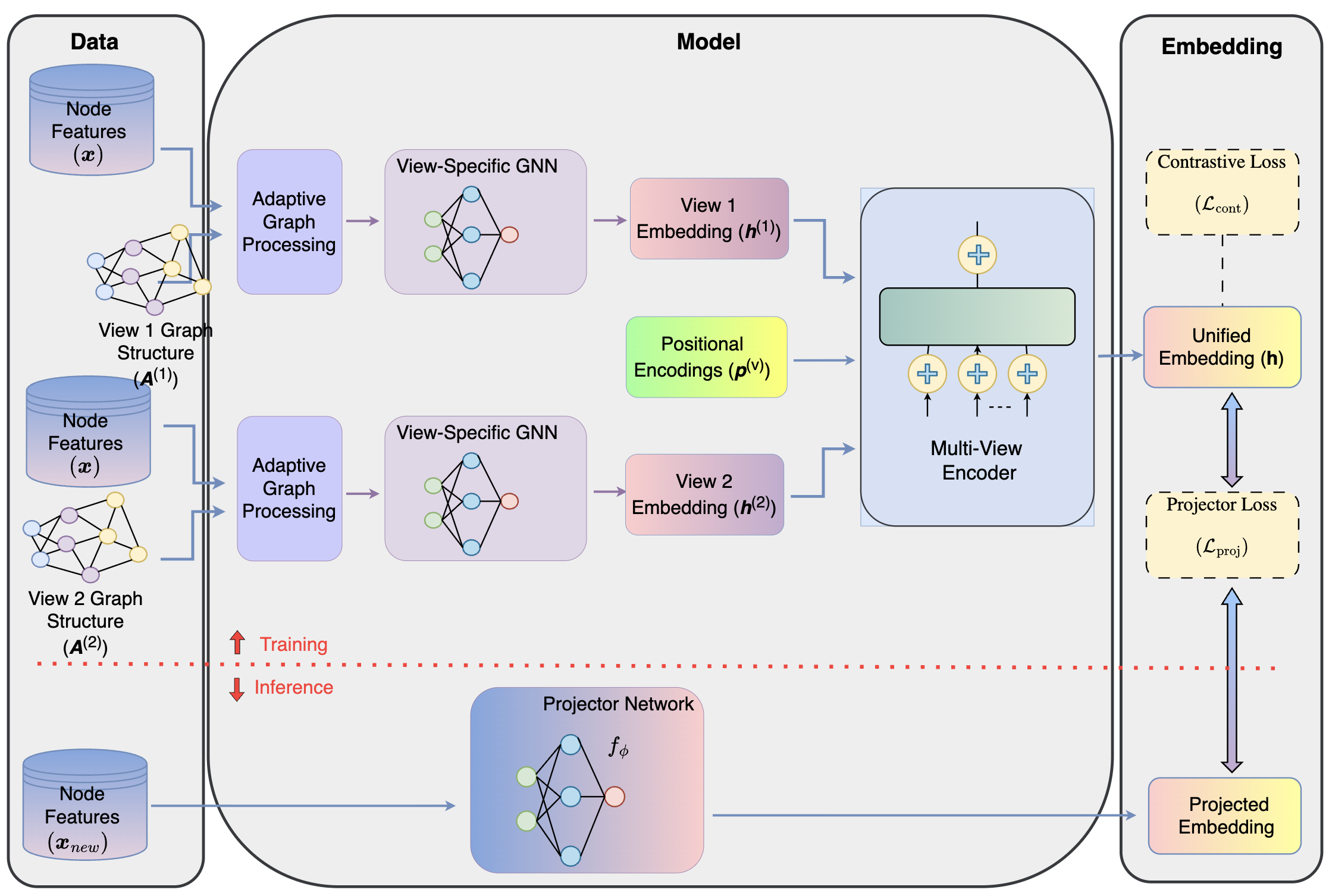}
\caption{The SNAP-KG framework~\cite{lin2026snapkg}. Adaptive graph preprocessing, view-specific
GNNs, and a multi-view encoder produce unified embeddings during
training; the projector $f_\phi$ is distilled to reproduce them from raw
features and is the only component used for streaming inference.}
\label{fig:pipeline}
\end{figure}

Figure~\ref{fig:pipeline} summarizes the framework. Homophily enters it
in three places. First, \emph{adaptive graph preprocessing} ranks each
node's neighbors by feature cosine similarity and keeps only the most
similar ones; under heterophily, the useful neighbors are the
dissimilar ones~\cite{zhu2020beyond}, so this step throws away the
structural signal. Second, the \emph{contrastive objective} treats a
node's kept neighbors as its positives and pulls their embeddings
together; under heterophily, this pulls nodes of different classes into
the same region of the embedding space. Third, the \emph{projector} is
distilled from those embeddings, so it inherits both effects: whatever
the teacher mixes together, the student cannot pull apart. None of this
is specific to SNAP-KG. It is the normal behavior of graph encoders
that assume homophily~\cite{zhu2020beyond}, and Section~\ref{sec:study}
shows that the transductive baselines exhibit the same performance drop.

\section{A Scope Study on Heterophilous Graphs}
\label{sec:study}

\paragraph{Setup.}
We evaluate on three graphs that are standard hard cases in the
heterophily literature: \textbf{Texas} and \textbf{Wisconsin}, webpage
networks from the WebKB collection~\cite{pei2020geom}, and
\textbf{Chameleon}, a Wikipedia page-link
network~\cite{rozemberczki2021multi} whose five classes come from
binning the page-traffic target. Each of these graphs is defined by a
single relation, so SNAP-KG operates on that one view. This raises an
obvious question: is the drop caused by heterophily, or simply by using
one view instead of several? Table~\ref{tab:oneview} separates the two.
All runs
use SNAP-KG's original settings, repeated using five random seeds. We report clustering
accuracy (ACC) and normalized mutual information (NMI); across seeds, the standard deviation of ACC stays below 3.7 points. For every dataset in this study, including those previously used to evaluate
SNAP-KG, we also report \emph{edge homophily}, the fraction of edges whose
two endpoints share a class label~\cite{zhu2020beyond}, computed per view on the raw graphs before adaptive filtering.

\begin{table}[t]
\centering
\caption{Edge homophily and SNAP-KG clustering quality across all eight
evaluated datasets. $h_{\max}$ / $h_{\mathrm{mean}}$: highest / mean
edge homophily across views. Upper block: benchmarks previously used to
evaluate SNAP-KG~\cite{lin2026snapkg}.
Lower block: new heterophilous results (mean over five random seeds).
The heterophilous graphs are single-relation and are run in SNAP-KG's
single-view configuration.}
\label{tab:homophily}
\begin{tabular}{lrrccrr}
\toprule
Dataset & Nodes & Views & $h_{\max}$ & $h_{\mathrm{mean}}$ & ACC (\%) & NMI (\%) \\
\midrule
ACM   & 3,025   & 2 & 0.82 & 0.73 & 91.50 & 71.08 \\
DBLP  & 4,057   & 3 & 0.80 & 0.60 & 91.82 & 74.03 \\
IMDB  & 4,780   & 2 & 0.62 & 0.55 & 46.67 & 3.42  \\
YELP  & 2,614   & 3 & 0.64 & 0.49 & 91.20 & 69.65 \\
MAG   & 113,919 & 2 & 0.66 & 0.66 & 68.20 & 50.85 \\
\midrule
Texas      & 183   & 1 & 0.06 & 0.06 & \texasACC{} & \texasNMI{} \\
Wisconsin  & 251   & 1 & 0.18 & 0.18 & \wiscACC{}  & \wiscNMI{}  \\
Chameleon  & 2,277 & 1 & 0.32 & 0.32 & \chamACC{}  & \chamNMI{}  \\
\bottomrule
\end{tabular}
\end{table}


\paragraph{Quality is associated with the availability of a homophilous view.}
Table~\ref{tab:homophily} places the new results next to the benchmarks
previously used to evaluate SNAP-KG~\cite{lin2026snapkg}. The pattern is clear, though it is a
correlation rather than proof of cause; Section~\ref{sec:nutshell} gives
the mechanism. Every dataset on which SNAP-KG achieves more than 90\% ACC has at
least one view with edge homophily of 0.64 or higher; MAG, at
$h_{\max}{=}0.66$, with four classes, and a much larger scale, still
reaches 68.20\%. The three heterophilous graphs have no view above 0.32 in terms of \emph{edge homophily},
and quality falls sharply: ACC drops to \texasACC{}\% on Texas,
\wiscACC{}\% on Wisconsin, and \chamACC{}\% on Chameleon, with NMI in the
single digits. IMDB shows that a moderately homophilous view is not
enough on its own: prior SNAP-KG results show that its views carry
little signal that separates the classes for any method~\cite{lin2026snapkg}. The
heterophilous graphs show the opposite problem: no view is homophilous at
all, so multi-view fusion has nothing reliable to build on. A KG with several
genuinely distinct heterophilous relations would also stress the fusion
stage; standard heterophily benchmarks do not provide a KG with several distinct heterophilous relations, so we leave that case to future work.

\begin{table}[t]
\centering
\caption{One homophilous and one heterophilous relation of DBLP and YELP,
each used on its own, from the original SNAP-KG single-view
ablation~\cite{lin2026snapkg}. Every row uses a single view, so the number of
views is fixed and only the relation changes.}
\label{tab:oneview}
\begin{tabular}{llrccrr}
\toprule
Dataset & Relation used & $h$ & ACC (\%) & NMI (\%) \\
\midrule
DBLP & APVPA  & 0.67 & 90.07 & 71.12 \\
DBLP & APTPA  & 0.32 & 56.42 & 21.51 \\
YELP & BSB    & 0.64 & 90.05 & 66.58 \\
YELP & BUB    & 0.45 & 37.18 & 0.57  \\
\bottomrule
\end{tabular}
\end{table}

\paragraph{Heterophily, not the number of views.}
Table~\ref{tab:oneview} fixes the number of views at one and lets only
the relation change, using results from the SNAP-KG single-view
ablation~\cite{lin2026snapkg}. One homophilous relation is already enough:
APVPA alone yields DBLP 90.07\% ACC against 91.82\% for the full
three-view model, while BSB alone yields YELP 90.05\% against 91.20\%. 
Heterophilous relations perform substantially worse: APTPA alone yields 56.42\% ACC on DBLP, while BUB alone yields 37.18\% on YELP, close to the 31.07–36.61\% range observed on Texas, Wisconsin, and Chameleon.
The drop therefore follows the homophily of the relation, not the number
of relations.

\begin{table}[t]
\centering
\caption{Clustering on the heterophilous graphs: SNAP-KG versus
transductive clustering baselines previously used to evaluate SNAP-KG.}
\label{tab:hetero}
\begin{tabular}{lcccccc}
\toprule
& \multicolumn{2}{c}{Texas} & \multicolumn{2}{c}{Wisconsin} & \multicolumn{2}{c}{Chameleon} \\
\cmidrule(lr){2-3}\cmidrule(lr){4-5}\cmidrule(lr){6-7}
Method & ACC (\%) & NMI (\%) & ACC (\%) & NMI (\%) & ACC (\%) & NMI (\%) \\
\midrule
AGE~\cite{cui2020adaptive}    & 33.99 & 6.14  & 33.55 & 10.47 & 36.68 & 10.59 \\
O2MAC~\cite{fan2020one2multi} & 49.18 & 4.10  & 45.82 & 5.67  & 32.15 & 8.95  \\
BMGC~\cite{shen2024balanced}  & 36.50 & 7.88  & 42.48 & 14.99 & 31.86 & 10.64 \\
DuaLGR~\cite{ling2023dual}    & 55.74 & 32.13 & 58.57 & 42.07 & 41.24 & 18.22 \\
SNAP-KG (ours)                & \texasACC{} & \texasNMI{} & \wiscACC{} & \wiscNMI{} & \chamACC{} & \chamNMI{} \\
\bottomrule
\end{tabular}
\end{table}

\paragraph{The limitation is shared, not specific.}
Table~\ref{tab:hetero} reports the transductive baselines from the
accompanying paper's evaluation on the same three graphs. Methods that
reach 82--93\% ACC on ACM and DBLP~\cite{lin2026snapkg} drop to
32--59\% here. DuaLGR drops the least, yet its best result, 58.57\% ACC
on Wisconsin, still trails the three benchmarks above
90\% by more than 30 points. Although SNAP-KG experiences a similar performance loss, it is the only method in the table that can assign
streaming entities inductively, without retraining. So the limitation
belongs to the whole family of homophily-assuming methods, transductive
and inductive alike, and not to any single method in it.

\section{Heterophilous Clustering Is a Different Problem}

Supporting heterophilous graphs is more than a minor modification to SNAP-KG; it is a
research direction of its own~\cite{gong2026survey}. Recent work
redesigns the representation itself: NGCE uses node features to guide
encoding and combines homophilous and heterophilous patterns in
multi-view graph clustering~\cite{ngce2025}; HeNCler learns an
asymmetric similarity for node clustering in heterophilous
graphs~\cite{hencler2025}; SMHGC mines similarity to strengthen
homophily for multi-view heterophilous clustering~\cite{sehg2024}. To the
best of our knowledge, these methods all target the transductive setting. This points to two independent axes. SNAP-KG provides \emph{inductive,
retraining-free deployment}; the works above provide
\emph{heterophily-aware representation}. No current method offers both.
Combining them is a concrete opportunity, because SNAP-KG's distillation
stage does not care how the teacher forms its embeddings: it only needs
feature-embedding pairs. Replacing the homophily-assuming GNN-Transformer
teacher with a heterophily-aware multi-view teacher, and distilling it
into the same projector $f_\phi$, would extend streaming entity
integration to heterophilous KGs without changing the deployment path. Until then, practitioners can tell in advance whether SNAP-KG is likely to work on a new KG by measuring the edge homophily of each relation, using a small labeled sample when one is available, or a feature-similarity estimate if they do not. If none of the relations is at least moderately homophilous, the resulting clusters should not be trusted.

\section{Conclusion}

This paper extends SNAP-KG~\cite{lin2026snapkg} with a scope study on heterophilous graphs. New five-seed SNAP-KG experiments on Texas,
Wisconsin, and Chameleon, together with a homophily measurement of all
eight datasets, show that SNAP-KG's strong results go together with
having at least one homophilous view, and that quality falls sharply when
none is present, both for SNAP-KG and for the transductive baselines. A
single-view comparison shows that this depends on the homophily of the
relation rather than on how many relations a dataset has.
This failure marks the boundary between two research problems: streaming
inductive deployment, which SNAP-KG solves, and heterophily-aware
representation, which a dedicated line of work
addresses~\cite{ngce2025,hencler2025,sehg2024}. Distilling a
heterophily-aware teacher into SNAP-KG's streaming projector is a
concrete path to serving both settings, and is the next direction for this work.

\newpage

\section*{Declaration on Generative AI}
We used a large language model (Claude, Anthropic) solely to assist
with grammar correction and writing refinement. All ideas, problem formulation,
methodology, experimental design, results, and conclusions are entirely 
our own. We take full responsibility for the accuracy and
integrity of all content presented in this work.

\bibliography{poster}

\end{document}